# A Note on Ontology and Ordinary Language

Walid S. Saba

*Computing and Statistical Sciences Center*
*American Institutes for Research*
*1000 Thomas Jefferson Street NW*
*Washington, DC 20007 USA*
*wsaba@air.org*

**Abstract.** We argue for a compositional semantics grounded in a strongly typed ontology that reflects our commonsense view of the world and the way we talk about it. Assuming such a structure we show that the semantics of various natural language phenomena may become nearly trivial.

## 1 Introduction

We argue that challenges in the semantics of natural language are rampant due to the gross mismatch between the trivial ontological commitments of our semantic formalisms and the reality of the world these formalisms purport to represent. In particular, we argue that semantics must be grounded in a much richer ontological structure, one that reflects our commonsense view of the world and the way we talk about it in ordinary language.

Recently, it was suggested in Saba (2007) how language itself could be used as a tool to discover (rather than invent) the nature of this ontological structure. The purpose of the current paper is to demonstrate that semantics could become 'nearly' trivial when grounded in such an ontological structure and this done by assuming the existence of a fairly simple and uncontroversial ontological structure. Furthermore, it will also be demonstrated here that it is the process of the semantic analysis itself that will in turn help us shed some light on the nature of this ontological structure, a structure that must be isomorphic to our commonsense view of the world and the way we talk about it in ordinary language.

## 2 Semantics with Ontological Content

We begin by making a case for a semantics that is grounded in a strongly typed ontological structure that is isomorphic to our commonsense view of

---

reality. In doing so, our ontological commitments will initially be minimal. In particular, we assume the existence of a subsumption hierarchy of a number of general categories such as `animal`, `substance`, `entity`, `artifact`, etc.

We shall use $(x :: \texttt{animal})$ to state that $x$ is an object of type `animal`, and $Articulate(x :: \texttt{human})$ to state that the property *Articulate* is true of some object $x$, an object that must be of type `human` (since 'articulate' is a property that is ordinarily said of humans). We write $(\exists x :: \texttt{t})(P(x))$ when the property $P$ is true of some object $x$ of type `t`; $(\exists^1 x :: \texttt{t})(P(x))$ when $P$ is true of a unique object of type `t`; and $(\exists x :: \texttt{t}^a)(P(x))$ when the property $P$ is true of some object $x$ of type `t`, an object that only conceptually (abstractly) exists - i.e., an object that need not physically exist. Proper nouns, such as *Sheba*, are interpreted as

(1) $[\![sheba]\!] \Rightarrow \lambda P[(\exists^1 x)(Noo(x :: \texttt{entity}, \text{'}sheba\text{'}) \wedge P(x :: \texttt{t}))]$,

where $Noo(x :: \texttt{entity}, s)$ is true of some individual object $x$ (of type `entity`), and $s$ if (the label) $s$ is the name of $x$. To simplify notation, we sometimes write (1) as $[\![sheba]\!] = \lambda P[(\exists^1 sheba :: \texttt{entity})(P(x :: \texttt{t}))]$. Let us define $Is(x,y)$ to be a predicate that is true of some $x$ and $y$ when $x$ is identical to $y$. Consider now the following:

(2) $[\![\textit{William H. Bonney is Billy the Kid}]\!]$
$\Rightarrow (\exists^1 x :: \texttt{entity})(\exists^1 y :: \texttt{entity})(Noo(x, \text{'}whb\text{'}) \wedge Noo(y, \text{'}btk\text{'}) \wedge Is(x,y))$
$\Rightarrow (\exists^1 whb :: \texttt{entity})(\exists^1 btk :: \texttt{entity})(Is(whb, btk))$

$[\![\textit{William H. Bonney is William H. Bonney}]\!]$
$\Rightarrow (\exists^1 x :: \texttt{entity})(\exists^1 y :: \texttt{entity})(Noo(x, \text{'}whb\text{'}) \wedge Noo(x, \text{'}whb\text{'}) \wedge Is(x,y))$
$\Rightarrow (\exists^1 whb :: \texttt{entity})(\exists^1 whb :: \texttt{entity})(Is(whb, whb))$
$\Rightarrow (\exists^1 whb :: \texttt{entity})(True()) \equiv (\exists^1 x :: \texttt{entity})(Noo(x, \text{'}whb\text{'}))$

This does seem plausible since '*William H. Bonney is Billy the Kid*' should have more content than '*William H. Bonney is William H. Bonney*' since the latter seems to only reiterate the existence of some '*whb*'[1].

Regarding associating types with variables it should be noted now that a variable might, in a single scope, be associated with more than one type. For example, $x$ in (1) is considered to be an `entity` and an object of type `t`, where `t` is presumably the type of objects that the property $P$ applies to (or makes sense of). In these situations some sort of type unification must occur, where the simplest case of type unification $(\texttt{s} \bullet \texttt{t})$, between two types $\texttt{s}$ and $\texttt{t}$, is defined as follows:

$$(3) \quad (\texttt{s} \bullet \texttt{t}) = \begin{cases} \texttt{s}, & \textit{if } (\texttt{s} \sqsubseteq \texttt{t}) \\ \texttt{t}, & \textit{if } (\texttt{t} \sqsubseteq \texttt{s}) \\ \bot, & \textit{otherwise} \end{cases}$$

[1] Note that $(\exists x :: \texttt{s})(\exists y :: \texttt{t})(P(x) \wedge Is(x,y)) \equiv (\exists x :: (\texttt{s} \bullet \texttt{t}))(P(x)) \equiv (\exists y :: (\texttt{s} \bullet \texttt{t}))(P(y))$.

To illustrate the notion of type unification, consider the steps involved in the interpretation of *sheba is hungry*, where we have assumed $(\texttt{animal} \sqsubseteq \texttt{entity})$, and that *Hungry* is a property that applies to (or makes sense of) objects that are of type $\texttt{animal}$.

$$\begin{aligned}
&[\![\textit{sheba is hungry}]\!] \\
&\Rightarrow (\exists^1 \textit{sheba} :: \texttt{entity})(\textit{Hungry}(\textit{sheba} :: \texttt{animal})) \\
&\Rightarrow (\exists^1 \textit{sheba} :: (\texttt{animal} \bullet \texttt{entity}))(\textit{Hungry}(\textit{sheba})) \\
&\Rightarrow (\exists^1 \textit{sheba} :: \texttt{animal})(\textit{Hungry}(\textit{sheba}))
\end{aligned}$$

Thus, *sheba is hungry* states that there is a unique object named *sheba*, which must be an object of type $\texttt{animal}$, and such that *sheba* is *Hungry*. Type unification will not always be as straightforward, and this will be discussed in some detail below. For now, however, we are interested in highlighting the utility of 'embedding' ontological sorts into the properties and relations of our logical forms. Consider for example the steps involved in the interpretation of *sheba is a young artist*, given in (4).

$$\begin{aligned}
(4) \quad &[\![\textit{sheba is a young artist}]\!] \\
&\Rightarrow (\exists^1 \textit{sheba} :: \texttt{entity})(\textit{Artist}(\textit{sheba} :: \texttt{human}) \wedge \textit{Young}(\textit{sheba} :: \texttt{physical})) \\
&\Rightarrow (\exists^1 \textit{sheba} :: (\texttt{entity} \bullet \texttt{physical}))(\textit{Artist}(\textit{sheba} :: \texttt{human}) \wedge \textit{Young}(\textit{sheba})) \\
&\Rightarrow (\exists^1 \textit{sheba} :: \texttt{physical})(\textit{Artist}(\textit{sheba} :: \texttt{human}) \wedge \textit{Young}(\textit{sheba})) \\
&\Rightarrow (\exists^1 \textit{sheba} :: (\texttt{human} \bullet \texttt{physical}))(\textit{Artist}(\textit{sheba}) \wedge \textit{Young}(\textit{sheba})) \\
&\Rightarrow (\exists^1 \textit{sheba} :: \texttt{human})(\textit{Artist}(\textit{sheba}) \wedge \textit{Young}(\textit{sheba}))
\end{aligned}$$

In the final analysis, therefore, '*sheba is a young artist*' is interpreted as follows: there is a unique object named *sheba*, an object of type $\texttt{human}$, and such that *sheba* is *Artist* and *Young*[2]. Note here that in contrast with $\texttt{human}$, which is a first-intension ontological concept (Cocchiarella, 2001), *Artist* and

---

[2] The type unifications in (4) can occur in any order since $(\texttt{r} \bullet (\texttt{s} \bullet \texttt{t})) = ((\texttt{r} \bullet \texttt{s}) \bullet \texttt{t})$. That is, type unification is associative (and of course commutative), and this is a consequence of the fact that $(\texttt{r} \sqcap (\texttt{s} \sqcap \texttt{t})) = ((\texttt{r} \sqcap \texttt{s}) \sqcap \texttt{t})$, where $\sqcap$ is the least upper bound (*lub*) operator. What is important, however, is that the type associated with the variable introduced by every quantifier be unified with the type of every property and relation, as demonstrated by later examples.

*Young* are considered to be second-intension logical concepts, namely properties that may or may not be true of first-intension (ontological) concepts[3]. Moreover, and unlike first-intension ontological concepts (such as `human`), logical concepts such as *Artist* and *Young* are assumed to be defined by virtue of logical expressions,

$$(\forall x :: \texttt{human})(Artist(x) \equiv_{df} \varphi_1) \text{ and}$$
$$(\forall x :: \texttt{animal})(Young(x) \equiv_{df} \varphi_2),$$

where the exact nature of $\varphi_1$ and $\varphi_2$ might very well be susceptible to temporal, cultural, and other contextual factors, depending on what, at a certain point in time, a certain community considers an *Artist* (for example) to be. That is, while the properties of being an *Artist* and *Young* that $x$ exhibits are accidental (as well as temporal, cultural-dependent, etc.), the fact that some $x$ is `human` (and thus an `animal`, etc.) is not[4].

# 3 More on Type Unification

Thus far we have performed simple type unifications involving types that are in a subsumption relationship. For example, we have suggested above that $(\texttt{human} \bullet \texttt{entity}) = \texttt{human}$, since $(\texttt{human} \sqsubseteq \texttt{entity})$, i.e., since a `human` is also an `entity`. Quite often, however, it is not subsumption but some other relationship that exists between the different types associated with a variable, and a typical example is the case of nominal compounds. In particular, we are interested in answering the question of what types of objects do the following nominal compounds, for example, refer to:

[3] Not recognizing the ontological difference between `human` and *Professor* (namely, that what ontologically exist are objects of type `human`, and not professors, and that *Professor* is a mere property that may or may not apply to objects of type `human`) has traditionally led to ontologies rampant with multiple inheritance.

[4] In a recent argument *Against Fantology,* Smith (2005) notes that too much attention has been paid to the false doctrine that much can be discovered about the ontological structure of reality by predication in first-order logic. According to Smith, for example, the use of standard predication in first-order logic in representing the meanings of '*John is a human*' and '*John is tall*' completely ignores the different ontological categories implicit in each utterance. While we agree with this observation, we believe that our approach to a semantics grounded in an a rich ontological structure that is supposed to reflect our commonsense reality, does solve this problem without introducing ad-hoc relations to the formalism, as example (4) and subsequent examples in this paper demonstrate. First-order logic (and Frege, for that matter), are therefore not necessarily the villains, and the ''predicates do not represent'' slogan is perhaps appropriate, but it seems only when predicates are devoid of any ontological content.

(5) a. *book review*
b. *book proposal*
c. *design review*
d. *design plan*

From the standpoint of commonsense, the existence of a *book* `review` should imply the existence of a `book`, while the existence of a *book* `proposal` should not (although it might after all exist, if, for example, we were speaking of a book proposal years after the publication of the book). Similar arguments can be made about the nominal compounds in (6c) and (6d)[5]. We could say therefore that a reference to a *book review* is a reference to a `review` (which is, ultimately, an `activity`), and the object of this activity must be an existing `book`; while a reference to a *book proposal* is a reference to a `proposal` of some `book`, a book that might not (yet) actually exist. That is,

(6) $[\![a\ \ book\ review]\!]$
$\Rightarrow \lambda P[(\exists x :: \texttt{book})(\exists y :: \texttt{review})(\mathit{ReviewOf}(y,x) \wedge P(y))]$
(7) $[\![a\ \ book\ proposal]\!]$
$\Rightarrow \lambda P[(\exists x :: \texttt{book}^a)(\exists y :: \texttt{proposal})(\mathit{ProposalFor}(y,x) \wedge P(y))]$

Note that $(Qx :: \texttt{t})(P(x)) \supset (Q^a x :: \texttt{t})(P(x))$, where $Q$ is one of the standard quantifiers $\forall$ and $\exists$ - that is, what actually exists must conceptually exist. Consequently, $(\forall x)(\forall \texttt{t})(x :: \texttt{t} \sqsubseteq x :: \texttt{t}^a)$ and according to our type unification rules $(\texttt{t}^a \bullet \texttt{t}) = \texttt{t}$. To summarize, type unification is finally defined as follows:

(8) $$(Qx :: (\texttt{s} \bullet \texttt{t}))(P(x)) \equiv \begin{cases} (Qx :: \texttt{s})(P(x)), & \mathit{if}\ (\texttt{s} \sqsubseteq \texttt{t}) \\ (Qx :: \texttt{t})(P(x)), & \mathit{if}\ (\texttt{t} \sqsubseteq \texttt{s}) \\ (Q^a x :: \texttt{s})(\exists y :: \texttt{t})(R(x,y) \wedge P(y)), & \mathit{if}\ (\exists R)(R(x,y)) \\ (Qx :: \bot)(P(x)), & \mathit{otherwise} \end{cases}$$

Finally, it must be noted that, in general, a type unification might fail, and this occurs in the absence of any relationship between the types assigned to a variable in the same scope. For example, assuming $\mathit{Artificial}(x :: \texttt{naturalObj})$, i.e., that *Artificial* is a property ordinarily said of objects of type `naturalObj`, and assuming $(\texttt{car} \sqsubseteq \texttt{artifact})$, then the nominal compound *artificial car* would get the interpretation

[5] In fact, it is precisely this kind of analysis that we are performing here that will help us shed some light on the nature of certain ontological categories, such as {`review`, `evaluation`, `analysis`, etc.} and {`proposal`, `suggestion`, `plan`, etc.}. In the appendix we suggest some template compositional functions for [*Noun Noun*] compounds involving a number of patterns.

(9) $[\![an\ \ artificial\ \ car]\!]$
$\Rightarrow \lambda P[(\exists x :: \texttt{car})(Artificial(x :: \texttt{naturalObj}))]$
$\Rightarrow \lambda P[(\exists x :: (\texttt{naturalObj} \bullet \texttt{car}))(Artificial(x))]$
$\Rightarrow \lambda P[(\exists x :: \bot)(Artificial(x))]$

It would seem therefore that type unification fails in the interpretation of some phrase that does not seem to be plausible from the standpoint of commonsense. It should also be noted here that there are nominal compounds that do not confirm with our commonsense (e.g., *former father*) that are not 'caught' with type unification, but are eventually caught at the logical level – See (Saba, 2007) for more details on this issue.

# 4 From Abstract to Actual Existence

Speaking of objects that might only conceptually exist, in addition to having a type in some assumed ontology, leads us to extend the notion of associating types with quantified variables in an important way.

Recall that our intention in associating types with quantified variables, as, for example, in $Articulate(x :: \texttt{human})$, was to reflect our commonsense understanding of how the property *Articulate* is used in our everyday discourse, namely that *Articulate* is ordinarily said of objects that are of type `human`. What of a property such as *Imminent*, then? Undoubtedly, saying some object $e$ is *Imminent* only makes sense in ordinary language when $e$ is some `event`, which we have been expressing as $Imminent(e :: \texttt{event})$. But there is obviously more that we can assume of $e$. In particular, *imminent* is said in ordinary language of some $e$ when $e$ is an `event` that has not yet occurred, that is, an event that exists only conceptually, which we write as $Imminent(e :: \texttt{event}^a)$. A question that arises now is this: what is the status of an event $e$ that, at the same time, is *imminent* as well as *important*? Clearly, an *important* and *imminent* `event` should still be assumed to be an event that does not actually exist (as important as it may be). Given our type unification rules, *important* must therefore be a property that is said of an event that also need not actually exist, as illustrated by the following:

(10) $[\![an\ \ important\ \ and\ \ imminent\ \ event]\!]$
$\Rightarrow \lambda P[(\exists x)(Importnat(x :: \texttt{entity}^a) \wedge Imminent(x :: \texttt{event}^a) \wedge P(x :: \texttt{t}))]$
$\Rightarrow \lambda P[(\exists x :: (\texttt{event}^a \bullet \texttt{entity}^a))(Importnat(x) \wedge Imminent(x) \wedge P(x :: \texttt{t}))]$
$\Rightarrow \lambda P[(\exists x :: \texttt{event}^a)(Importnat(x) \wedge Imminent(x) \wedge P(x :: \texttt{t}))]$

It is important to note here that one can always 'bring down' an object (such as an `event`) from abstract existence into actual existence, but the reverse is not true. Consequently, quantification over variables associated with the type of an abstract concept, such as `event`, should always initially assume abstract existence. To illustrate, let us first assume the following:

(11) $Attend(x :: \texttt{human}, y :: \texttt{event})$
$Cancel(x :: \texttt{human}, y :: \texttt{event}^{a})$

That is, we have assumed here that it always makes sense to speak of a `human` that attended or cancelled some `event`, where to attend an event is to have an existing event; and where the object of a cancellation is an `event` that does not (anymore, if it ever did) exist[6]. Consider now the following:

(12) $[\![ \textit{john attended the seminar} ]\!]$
$\Rightarrow (\exists^1 j :: \texttt{human})(\exists^1 e :: \texttt{seminar}^{a})(Attended(j :: \texttt{human}, e :: \texttt{event}))$
$\Rightarrow (\exists^1 j :: (\texttt{human} \bullet \texttt{human}))(\exists^1 e :: (\texttt{seminar}^{a} \bullet \texttt{event}))(Attended(j, e))$
$\Rightarrow (\exists^1 j :: \texttt{human})(\exists^1 e :: \texttt{seminar})(Attended(j, e))$

That is, saying '*john attended the seminar*' is saying there is a specific `human` named $j$, a specific `seminar` $e$ (that actually exists) such that $j$ attended $e$. On the other hand, consider now the interpretation of the sentence in (13).

(13) $[\![ \textit{john cancelled the seminar} ]\!]$
$\Rightarrow (\exists^1 john :: \texttt{human})(\exists^1 y :: \texttt{seminar})(Cancelled(john :: \texttt{human}, y :: \texttt{event}^{a}))$
$\Rightarrow (\exists^1 john :: (\texttt{human} \bullet \texttt{human}))(\exists^1 y :: (\texttt{seminar} \bullet \texttt{event}^{a}))$
$(Cancelled(john, y))$
$\Rightarrow (\exists^1 john :: \texttt{human})(\exists^1 y :: \texttt{seminar}^{a})(Cancelled(john, y))$

What (13) states is that there is a specific `human` named *john*, and a specific `seminar` (that does not necessarily exist), a seminar that *john* cancelled[7]. An interesting case now occurs when a type is 'brought down' from abstract existence into actual existence. Le us assume $Plan(x :: \texttt{human}, y :: \texttt{event}^{a})$; that

[6] Tense and modal aspects can also effect the initial type assignments. For example, in '*john will attend the seminar*' the initial assumption should be that the seminar (`event`) might not yet actually exist. While this does not affect the (different) argument being made here, a full treatment of this issue here would complicate the presentation considerably as this would involve discussing the interaction with syntax in much more detail.

[7] As Hirst (1991) correctly notes, assuming that the reference to the *seminar* is intensional, i.e., that the reference is to 'the idea of a seminar' does not solve the problem since the idea of a seminar is not what was cancelled, but *an actual event that did not actually happen*!

is, it always makes sense to say that some `human` is planning (or did plan) an `event` that need not (yet) actually exist. Consider now the following,

(14) $[\![\textit{john planned the trip}]\!]$
$\Rightarrow (\exists^1 j :: \mathtt{human})(\exists^1 e :: \mathtt{trip}^a)(\mathit{Planned}(x :: \mathtt{human}, y :: \mathtt{event}^a))$
$\Rightarrow (\exists^1 j :: (\mathtt{human} \bullet \mathtt{human}))(\exists^1 e :: (\mathtt{trip}^a \bullet \mathtt{event}^a))(\mathit{Planned}(j, e))$
$\Rightarrow (\exists^1 j :: \mathtt{human})(\exists^1 e :: \mathtt{trip}^a)(\mathit{Planned}(j, e))$

That is, saying *john planned the trip* is simply saying that a specific object that must be a `human` has planned a specific `trip`, a `trip` that might not have actually happened[8]. However, assuming $\mathit{Lengthy}(e :: \mathtt{event})$; i.e., that *Lengthy* is a property that is ordinarily said of an (existing) event, then the interpretation of '*john planned the lengthy trip*' should proceed as follows:

(15) $[\![\textit{john planned the lengthy trip}]\!]$
$\Rightarrow (\exists^1 j :: \mathtt{human})(\exists^1 e :: \mathtt{trip}^a)(\mathit{Planned}(x :: \mathtt{human}, y :: \mathtt{event}^a)$
$\quad \wedge \mathit{Lengthy}(e :: \mathtt{event}))$
$\Rightarrow (\exists^1 j :: \mathtt{human})(\exists^1 e :: \mathtt{trip}^a)(\mathit{Planned}(j, e :: (\mathtt{event} \bullet \mathtt{event}^a)) \wedge \mathit{Lengthy}(e))$
$\Rightarrow (\exists^1 j :: \mathtt{human})(\exists^1 e :: (\mathtt{trip}^a \bullet \mathtt{event}))(\mathit{Planned}(j, e) \wedge \mathit{Lengthy}(e))$
$\Rightarrow (\exists^1 j :: \mathtt{human})(\exists^1 e :: \mathtt{trip})(\mathit{Planned}(j, e) \wedge \mathit{Lengthy}(e))$

That is, there is a specific `human` (named *john*) that has planned a specific `trip`, a trip that was *Lengthy.* It should be noted here that the `trip` in (15) was finally considered to be an existing event due to other information contained in the same sentence. In general, however, this information can be contained in a larger discourse. For example, in interpreting

(16) *John planned the trip. It was lengthy.*

the resolution of 'it' would force a retraction of the types inferred in processing '*John planned the trip*', as the information that follows will 'bring down' the aforementioned trip from abstract to actual existence. Such details are clearly beyond the scope of this paper, but readers interested in the computational details of such processes are referred to (van Deemter & Peters, 1996).

[8] It is the trip (`event`) that did not necessarily happen, not the planning (`activity`) for it.

## 5 On Intensional Verbs and Dot (•) Objects

Consider the following sentences and their corresponding translation into standard first-order logic:

(17) $[\![\textit{john found a unicorn}]\!] \Rightarrow (\exists x)(Unicorn(x) \wedge Found(j, x))$
(18) $[\![\textit{john sought a unicorn}]\!] \Rightarrow (\exists x)(Unicorn(x) \wedge Sought(j, x))$

Note that $(\exists x)(Elephant(x))$ can be inferred in both cases, although it is clear that '*john sought a unicorn*' should not entail the existence of a unicorn. In addressing this problem, Montague (1960) suggested a solution that in effect treats 'seek' as an intensional verb that has more or less the meaning of 'tries to find', using the tools of a higher-order intensional logic. In addition to unnecessary complication of the logical form, however, we believe that this is, at best, a partial solution since the problem in our opinion is not necessarily in the verb *seek*, nor in the reference to unicorns. That is, *painting*, *imagining*, etc. of a unicorn (or an elephant, for that matter) should not entail the existence of a unicorn (nor the existence of an elephant). To illustrate further, let us first assume the following:

(19) $Paint(x :: \texttt{human}, y :: \texttt{painting})$
(20) $Find(x :: \texttt{human}, y :: \texttt{entity})$

That is, we are assuming that it always makes sense to speak of a `human` that painted some `painting`, and of some `human` that found some `entity`. Consider now the interpretation in (21), where it was assumed that *Large* is a property that applies to (or makes sense of) objects that are of type `physical`.[9]

(21) $[\![\textit{john found a large elephant}]\!]$

$$\Rightarrow (\exists^1 john :: \texttt{human})(\exists e :: \texttt{elephant})$$
$$(Found(j :: \texttt{human}, e :: \texttt{entity}) \wedge Large(e :: \texttt{physical}))$$
$$\Rightarrow (\exists^1 john :: (\texttt{human} \bullet \texttt{human}))(\exists e :: (\texttt{elephant} \bullet \texttt{physical}))$$
$$(Found(j, e :: \texttt{entity}) \wedge Large(e))$$
$$\Rightarrow (\exists^1 john :: \texttt{human})(\exists e :: \texttt{elephant})(Found(j, e :: \texttt{entity})) \wedge Large(e))$$
$$\Rightarrow (\exists^1 john :: \texttt{human})(\exists e :: (\texttt{elephant} \bullet \texttt{entity}))(Found(j, e)) \wedge Large(e))$$
$$\Rightarrow (\exists^1 john :: \texttt{human})(\exists e :: \texttt{elephant})(Found(j, e)) \wedge Large(e))$$

[9] Of course, we are also assuming here that $(\texttt{elephant} \sqsubseteq \texttt{physical} \sqsubseteq \texttt{entity})$.

In the final analysis, therefore, if '*john found a large elephant*' then there is a specific `human` (named *j*), and some `elephant` *e*, such that *e* is *Large* and *j* found *e*. However, consider now the interpretation in (22).

(22) $[\![\textit{john painted a large elephant}]\!]$
$\Rightarrow (\exists^1 john :: \texttt{human})(\exists e :: \texttt{elephant})$
$(Painted(j :: \texttt{human}, e :: \texttt{painting}) \wedge Large(e :: \texttt{physical}))$

Note that what we now have is a quantified variable, *e*, that is supposed to be an object of type `elephant`, an object that is described by a property, where it is considered to be an object of type `physical`, and an object that is in a relation in which it is considered to be a `painting`. In this case there are two pairs of type unifications that must occur, namely ($\texttt{elephant} \bullet \texttt{painting}$) and ($\texttt{elephant} \bullet \texttt{physical}$), where the former would result in the introduction of a new variable of type `painting`. This process, depicted graphically in figure 1 below, would in the final analysis result in the following:

(23) $[\![\textit{john painted a large elephant}]\!]$
$\Rightarrow (\exists^1 john :: \texttt{human})(\exists^a e :: \texttt{elephant})(\exists p :: \texttt{painting})$
$(Painted(j, p) \wedge PaintingOf(p, e) \wedge Large(e))$

Note here that the interpretation correctly states that it is a (painted) elephant (that need not actually exist) that is *Large* and not the painting itself. Thus, '*john painted an elephant*' is correctly interpreted as roughly meaning '*john made a painting of a large elephant*'[10].

In addition to handling the so-called intensional verbs, our approach seems to also appropriately handle other situations that, on the surface, seem to be addressing a different issue. For example, consider the following:

(24) *john read the book and then he burned it.*

In Asher and Pustejovsky (2005) it is argued that 'book' in this context must have what is called a dot type, which is a complex structure that in a sense carries within it the semantic types associated with various senses of 'book'. For instance, it is argued that 'book' in (24) carries the 'informational content' sense (when it is being read) as well as the 'physical object' sense (when it is being burned). Elaborate machinery is then introduced to 'pick out' the right sense in the right context, and all in a well-typed compositional logic. But this approach presupposes that one can enumerate, a priori, all possible uses of the word 'book' in ordinary language[11].

[10] To get this interpretation we must assume $PaintingOf(x :: \texttt{painting}, y :: \texttt{physical}^a)$, i.e., that we can always speak of a `painting` of some `physical` object that need not actually exist.

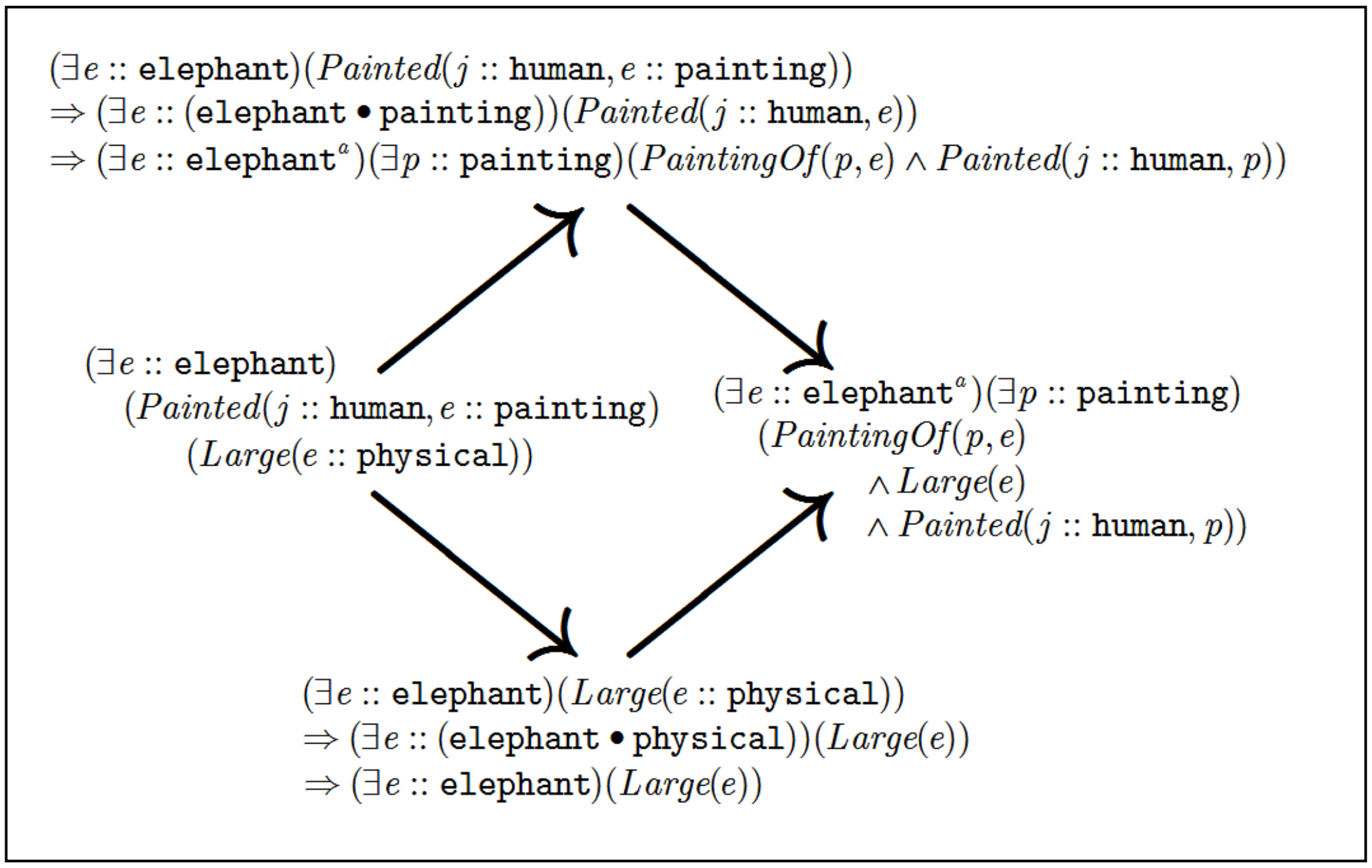

**Figure 1.** A pair of type unifications (that can happen in parallel).

Moreover, this approach does not seem to provide a solution for the problem posed by example (23), since there does not seem to be an obvious reason why a complex dot type for 'elephant' should contain a representational sense, although it is an object that can be painted. To see how this problem is dealt with in our approach, consider the following:

(24) $\mathit{Read}(x :: \texttt{human}, y :: \texttt{content})$
(25) $\mathit{Burn}(x :: \texttt{human}, y :: \texttt{physical})$

That is, we are assuming here that it always makes sense to speak of a `human` that read some `content`, and of a `human` that burned some `physical` object. Consider now the following:

---

[11] Similar presuppositions are also made in a hybrid (connectionist/symbolic) 'sense modulation' approach described by Rais-Ghasem and Corriveau (1998).

(26) $[\![john\ read\ a\ book]\!]$
$\Rightarrow (\exists^1 j :: \texttt{human})(\exists b :: \texttt{book})(Read(j :: \texttt{human}, b :: \texttt{content}))$
$\Rightarrow (\exists^1 j :: (\texttt{human} \bullet \texttt{human}))(\exists b :: (\texttt{book} \bullet \texttt{content}))(Read(j, b))$
$\Rightarrow (\exists^1 j :: \texttt{human})(\exists b :: \texttt{book})(\exists c :: \texttt{content})(ContentOf(b, c) \wedge Read(j, b))$

Thus, if '*john read a book*' then there is some specific `human` (named $j$), some object $b$ of type `book`, such that that $j$ read the content of $b$. On the other hand, consider now the following:

(27) $[\![john\ burned\ a\ book]\!]$
$\Rightarrow (\exists^1 j :: \texttt{human})(\exists b :: \texttt{book})(Burn(j :: \texttt{human}, b :: \texttt{physical}))$
$\Rightarrow (\exists^1 j :: (\texttt{human} \bullet \texttt{human}))(\exists b :: (\texttt{book} \bullet \texttt{physical}))(Burned(j, b))$
$\Rightarrow (\exists^1 j :: \texttt{human})(\exists b :: \texttt{book})(Burned(j, b))$

That is, if '*john burned a book*' then there is some specific `human` (named $j$), some object $b$ of type `book`, such that that $j$ burned $b$. Note, therefore, that when the `book` is being burned we are simply referring to the book as the `physical` object that it is, while reading the book implies, implicitly, that we are referring to an additional (abstract) object, namely the `content` of the book. The important point we wish to make here is that there is one `book` object, an object that is (ultimately) a `physical` object, that one can read (its `content`), sell/trade/etc (as a `commodity`), ..., or burn (as is, i.e., as simply the `physical` object that it is!) This means that 'book' can be easily used in different ways in the same linguistic context, as illustrated by the following:

(27) $[\![john\ read\ a\ book\ and\ then\ he\ burned\ it]\!]$
$\Rightarrow (\exists^1 j :: \texttt{human})(\exists b :: \texttt{book})(Read(j :: \texttt{human}, b :: \texttt{content})$
$\wedge\ ContentOf(b, c) \wedge Burn(j :: \texttt{human}, b :: \texttt{physical}))$

Like the example of 'painting a large elephant' discussed in (23) above, where the painting of an elephant implied its existence in some `painting` and it being large as some physical object (that need not actually exist), in (27) we also have a reference to a book as a `physical` object (that has been burned), and to a book that has `content` (that has been read). Similar to the process depicted in figure 1 above, the type unifications in (27) should now result in the following:

(28) $[\![john\ read\ a\ book\ and\ then\ he\ burned\ it]\!]$
$\Rightarrow (\exists^1 j :: \texttt{human})(\exists b :: \texttt{book})(\exists c :: \texttt{content})$
$(ContentOf(c, b) \wedge Read(j, c) \wedge Burn(j, b))$

That is, there is some unique object of type `human` (named $j$), some `book` $b$, some `content` $c$, such that $c$ is the `content` of $b$, and such that $j$ read $c$ and burned $b$. As pointed out in a previous section, it should also be noted here that these type unifications are often retracted in the presence of additional information. For example, in

(29) *John borrowed Das Kapital from the library. He did not agree with it.*

the resolution of 'it' would eventually result in the introduction of an (abstract) object of type `content` (which one might not agree with), as one does not agree (or disagree) with a `physical` object, an object that can indeed be borrowed[12].

# 6 All Variables were Created Equal

In this section we briefly discuss to the representation of various abstract types (such as events, properties, activities, etc.). First, and notwithstanding various extensions and modifications to Davidson's (1980) original theory, the advantages of treating events as individual objects that can be quantified over and described in various ways are, we believe, universally accepted.

However, there does not seem to be an obvious reason an `event` (in contrast with an `attribute`, a `property`, a `state`, a `process`, a `feeling`, etc.) should receive a special ontological status, and in particular, since we clearly treat such categories as predicable objects in ordinary language. For example, consider the following, where it is assumed that *Exhausting* is a property that is ordinarily said of events, i.e., $\mathit{Exhausting}(e :: \mathtt{event})$ and that $(\mathtt{activity} \sqsubseteq \mathtt{event})$:

(30) *John planned the trip. It was exhausting.*

In (30) 'it' can potentially refer to the trip (`event`), but it can also refer to the planning (`activity`). Thus an appropriate representation of (30) must have a reference to an object of type `activity`, and this can be done as follows:

(31) $[\![$*john planned the trip. It was exhausting.*$]\!]$

[12] Interestingly, in addition to introducing a `content` object, the resolution of 'it' would trivially result in *Das Kapital*, since you cannot also agree or disagree with a `library`, but, again, with the content of the library's books.

$\Rightarrow (\exists^1 j :: \texttt{human})(\exists^1 a :: \texttt{activity})(\exists^1 e :: \texttt{trip})$
$\quad (Planning(a) \wedge Subject(a, j) \wedge Object(a, e) \wedge Exhausting(a))$
$\Rightarrow (\exists^1 j :: \texttt{human})(\exists^1 a :: \texttt{activity})(\exists^1 e :: \texttt{trip})$
$\quad (Planning(a) \wedge Subject(a, j) \wedge Object(a, e) \wedge Exhausting(e))$

That is, there is a specific `human` (named *j*), a specific (*planning*) `activity` *a*, and a specific (`event`) *e*, such that *j* performed *a*, *e* was the object *a*, and such that either *a* or *e* was *Exhausting.* To highlight the fact that an abstract object (such as an `event`, `attribute`, `property`, `state`, `process`, etc.) should be treated like any other object, consider also the following:

(30) a. *Sheba is hungry*
b. *Running is fun*
c. *Nobility is desirable*
d. *Aging is inevitable*

Much like '*sheba'* has no instances, but is in fact the name of some instance of type human, there also no instances of 'nobility', and 'nobility' is simply the name of a specific `attribute`; and similarly, '*running*' is the name of some `activity`; and '*aging*' is the name of some `process`, etc., which could be expressed as follows:

(31) a. $[\![sheba\ is\ hungry]\!] \Rightarrow (\exists^1 x :: \texttt{human})(Noo(x, \text{`}sheba\text{'}) \wedge Hungry(x))$
b. $[\![running\ is\ fun]\!] \Rightarrow (\exists^1 a :: \texttt{activity})(Noo(a, \text{`}running\text{'}) \wedge Fun(a))$
c. $[\![nobility\ is\ desirbale]\!] \Rightarrow (\exists^1 a :: \texttt{attribute})(Noo(a, \text{`}nobility\text{'}) \wedge Desirbale(a))$
d. $[\![aging\ is\ inevitable]\!] \Rightarrow (\exists^1 p :: \texttt{process})(Noo(a, \text{`}againg\text{'}) \wedge Inevitable(a))$

That is, while (31a) is a statement about some individual object, namely that a `human` named *sheba* is in some `state`, the rest of the sentences can be read as follows: an `activity` named '*running*' is fun (31b); an `attribute` named '*nobility*' is desirable (31c); and a `process` named '*aging*' is inevitable (31d). In this regard, the representation we are suggesting here seems to also resolve the debate regarding the traditional difference between the 'is' of identity and the 'is' of predication. To illustrate, let us again consider the following:

(33) $[\![William\ H.\ Bonney\ is\ Billy\ the\ Kid\ ]\!]$
$\Rightarrow (\exists^1 x :: \texttt{entity})(\exists^1 y :: \texttt{entity})(Noo(x, \text{`}whb\text{'}) \wedge Noo(y, \text{`}btk\text{'}) \wedge Is(whb, btk))$
$\Rightarrow (\exists^1 whb :: \texttt{entity})(\exists^1 btk :: \texttt{entity})(Is(whb, btk))$

That is, there is a specific `entity` named *whb* and a specific `entity` named *btk*, and *whb* is (identical to) *btk* (note that in the absence of any additional information all that can be said of the objects in (33) is that they are objects of type `entity`). The use of 'is' in the context of sentences such as (33) is generally considered to be the 'is' of identity. However, consider now the following:

(34) $[\![\textit{William H. Bonney is a thief}\,]\!]$
$\Rightarrow (\exists^1 whb :: \texttt{entity})(\exists x :: \texttt{human})(\mathit{Thief}(x) \wedge \mathit{Is}(whb, x))$
$\Rightarrow (\exists^1 whb :: (\texttt{human} \bullet \texttt{entity}))(\exists x)(\mathit{Thief}(x) \wedge \mathit{Is}(whb, x))$
$\Rightarrow (\exists^1 whb :: \texttt{human})(\exists x)(\mathit{Thief}(x) \wedge \mathit{Is}(whb, x))$

That is, '*William H. Bonney is a theif*' is initially interpreted as follows: there is a unique `entity`, named *whb*, some object $x$ of type `human`, such that $x$ is a *Thief*, and such that *whb* is (identical to) $x$. However, since $x$ is identical to ' *whb*' (34) the variable can be removed, resulting in the following:

(34) $[\![\textit{William H. Bonney is a thief}\,]\!]$
$\Rightarrow (\exists^1 whb :: \texttt{human})(\exists x)(\mathit{Thief}(x) \wedge \mathit{Is}(whb, x))$
$\Rightarrow (\exists^1 whb :: \texttt{human})(\mathit{Thief}(whb))$

The same result is also obtained when interpreting sentences such as '*john is young*' and '*john is running*' since these sentences essentially mean '*john is a young thing*' and '*john is a running thing*', respectively.

# 7 Discussion

If the main business of semantics is to explain how linguistic constructs relate to the world, then semantic analysis of natural language text is indirectly an attempt at uncovering the semiotic ontology of commonsense knowledge, and particularly the background knowledge that seems to be implicit in all that we say in our everyday discourse. While this intimate relationship between language and the world is generally accepted, semantics (in all its paradigms) has traditionally proceeded in one direction: by first stipulating an assumed set of ontological commitments followed by some machinery that is supposed to, somehow, model meanings in terms of that stipulated structure of reality.

Given the gross mismatch between ordinary language and some of these presupposed ontological commitments, it is not surprising that difficulties in the semantic analysis of various natural language phenomena are rampant. As Hobbs (1985) correctly observed some time ago, however, semantics could become nearly trivial if it was grounded in an ontological structure that is "isomorphic to the way we talk about the world", as we also tried to demonstrate in this paper. However, a valid question that one might ask now is the following: how does one arrive at this ontological structure that implicitly underlies all that we say in everyday discourse? One plausible answer is the (seemingly circular) suggestion that the semantic analysis of natural language should itself be used to uncover this structure. In this regard we strongly agree with Dummett (1991) who states:

> We must not try to resolve the metaphysical questions first, and then construct a meaning-theory in light of the answers. We should investigate how our language actually functions, and how we can construct a workable systematic description of how it functions; the answers to those questions will then determine the answers to the metaphysical ones.

What this suggests, and correctly so, in our opinion, is that in our effort to understand the complex and intimate relationship between ordinary language and everyday (commonsense) knowledge, one could, as Bateman (1995) has also suggested, "use language as a tool for uncovering the semiotic ontology of commonsense" since language is the only theory we have of everyday knowledge. To alleviate this seeming circularity in wanting this ontological structure that would trivialize semantics; while at the same time suggesting that semantic analysis of language should itself be used to uncover this ontological structure, we suggested in this paper performing semantic analysis from the ground up, assuming a minimal (almost a trivial and basic) ontology, building up the ontology as we go guided by the results of the semantic analysis. The advantages of this approach are: (*i*) the ontology thus constructed as a result of this process would not be *invented*, as is the case in most approaches to ontology (e.g., Guarino, 1995, Lenat and Guha, 1990, and Sowa, 1995), but would instead be *discovered* from what is in fact implicitly assumed in our use of language in everyday discourse; (*ii*) the semantics of several natural language phenomena should as a result become trivial, since the semantic analysis was itself the source of the underlying knowledge structures (in a sense, one could say that the semantics would have been done before we even started!)

Finally it should be noted that we would certainly have plenty of work left even if semantics became nearly trivial when grounded in an ontological structure that is isomorphic to the world and the way we talk about it, as there is much difficult work left to be done at the discourse/pragmatic level.

# References

Asher, N. and Pustejovsky, J. (2005), Word Meaning and Commonsense Metaphysics, available from semanticsarchive.net

Bateman, J. A. (1995), On the Relationship between Ontology Construction and Natural Language: A Socio-Semiotic View, *International Journal of Human-Computer Studies*, **43**, pp. 929-944.

Cocchiarella, N. B. (2001), Logic and Ontology, *Axiomathes*, **12**, pp. 117-150.

Davidson, D. (1980), *Essays on Actions and Events*, Oxford: Clarendon Press.

Dummett. M. (1991), *The Logical Basis of Metaphysics*, Duckworth, London.

Guarino, N. (1955), Formal Ontology in Conceptual Analysis and Knowledge Representation, *International Journal of Human-Computer Studies*, **43** (5/6), Academic Press.

Hirst, G. (1991), Existence Assumptions in Knowledge Representation, *Artificial Intelligence*, **49** (3), pp. 199-242.

Hobbs, J. (1985), Ontological Promiscuity, In *Proc. of the 23rd Annual Meeting of the Assoc. for Computational Linguistics*, pp. 61-69, Chicago, Illinois, 1985.

Lenat, D. B., Guha, R. V., 1990. *Building Large Knowledge-Based Systems: Representation & Inference in the CYC Project.* Addison-Wesley.

Montague, R. (1960), On the Nature of certain Philosophical Entities, *The Monist*, **53**, pp. 159-194.

Rais-Ghasem, M. and Corriveua, J.-P. (1998), Exampler-Based Sense Modulation, In *Proceedings of COLING-ACL '98 Workshop on The Computational Treatment of Nominals.*

Saba, W. S. (2007), Language, Logic and Ontology - Uncovering the Structure of Commonsense Knowledge, *International Journal of Human-Computer Studies*, in press, doi:10.1016/j.ijhcs.2007.02.002.

Smith, B. (2005), Against Fantology, In M. E. Reicher and J. C. Marek (Eds.), *Experience and Analysis*, pp. 153-170, Vienna: HPT&OBV.

Sowa, J.F., 1995. Knowledge Representation: Logical Philosophical, and Computational Foundations. PWS Publishing Company, Boston.

van Deemter, K., Peters, S. 1996. (Eds.), *Semantic Ambiguity and Underspecification.* CSLI, Stanford, CA

# Appendix

In section 3 we suggested the following interpretations involving the nominal compounds '*a book review*' and '*a book proposal*':

(1) $[\![a\ \ book\ review]\!] \Rightarrow \lambda P[(\exists x :: \texttt{book})(\exists y :: \texttt{review})(ReviewOf(y,x) \wedge P(y))]$
(2) $[\![a\ \ book\ proposal]\!] \Rightarrow \lambda P[(\exists x :: \texttt{book}^{a})(\exists y :: \texttt{proposal})(ProposalFor(y,x) \wedge P(y))]$

In fact it is this kind of analysis itself that seems to shed some light on the nature of these ontological categories. For example, we suggest that the following compositional function is a template for all [*Noun Noun*] compounds shown in table 1 below.

(3) $[\![a\ N_{\texttt{substance}}\ \ N_{\texttt{artifact}}]\!]$
$\Rightarrow \lambda P[(\exists x :: \texttt{artifact})(\exists y :: \texttt{substance})(MadeOf(x,y) \wedge P(x))]$

Note, further, that the '*MadeOf*' relation seems to be specialized for specific types of `substance` and `artifact`. That is, while we *build* a `house`, we *erect* a `statue`, *knit* a shirt, *prepare* a `salad`, *bake* a `cake`, etc. Thus, building, erecting, knitting, baking, etc. are all different (senses) ways of making, and this exactly why the verb 'make' is highly polysemous.

| | | |
|---|---|---|
| *brick house* | *silk tie* | *rice pudding* |
| *silver spoon* | *cotton shirt* | *cheese cake* |
| *paper cup* | *leather boots* | *ham sandwich* |
| *plastic knife* | *wool sweater* | *fruit salad* |
| *marble statue* | *denim jeans* | *orange juice* |
| (*a*) | (*b*) | (*c*) |

**Table 1.** Patterns of [*Noun Noun*] nominal compounds

What would be interesting here is to be able to find out all of the generic compositional functions that are needed for an adequate treatment of all nominal compounds.

Finally, it should be noted that the same seems to also be true in the case of [*Adj Noun*] compounds. For example, it seems that for an object $x$, which must be of type `human`, *Former P*, where $P$ is a property such as *president*, *coach*, *senator*, etc., has the following interpretation:

(4) $[\![a\ Former\ P]\!]$
$\Rightarrow \lambda P[(\exists x :: \texttt{human})(\exists t)((t < t_u) \wedge [\![P]\!](x,t) \wedge \neg[\![P]\!](x, now))]$

where $t_u$ is the time of utterance. That is, some object $x$, which must be of type **human**, is a *Former P* if $x$ was (at some point in the past), and is not now a *P*. Note also that while '*former*' combines with temporal role types according to (4), but not with roles such as *father*, *doctor*, etc.